# Structured Arc Reversal and Simulation of Dynamic Probabilistic Networks

**Adrian Y. W. Cheuk and Craig Boutilier**
Department of Computer Science
University of British Columbia
Vancouver, BC, CANADA, V6T 1Z4
**email:** cheuk,cebly@cs.ubc.ca

## Abstract

We present an algorithm for arc reversal in Bayesian networks with tree-structured conditional probability tables, and consider some of its advantages, especially for the simulation of dynamic probabilistic networks. In particular, the method allows one to produce CPTs for nodes involved in the reversal that exploit regularities in the conditional distributions. We argue that this approach alleviates some of the overhead associated with arc reversal, plays an important role in evidence integration and can be used to restrict sampling of variables in DPNs. We also provide an algorithm that detects the *dynamic irrelevance* of state variables in forward simulation. This algorithm exploits the structured CPTs in a reversed network to determine, in a time-independent fashion, the conditions under which a variable does or does not need to be sampled.

## 1 Introduction

Recent investigations have explored the extension of the types of independence that can be represented in Bayesian networks (BNs). Specifically, the conditional independence of variables given a certain *context* (or instantiation of variables) has been proposed as a way of making BN specification and inference more tractable [8, 15, 2]. This *context-specific independence* (CSI) is often represented by the use of structured representations of the conditional probability tables (CPTs) for the network. While a variable is directly dependent on all of its parents, structured CPT representations, such as decision trees [2] or rules [15], capture the fact that (direct) dependence on certain parents does not hold given particular instantiations of others. The development of algorithms that exploit CSI, and the integration of CSI with techniques for manipulating BNs and influence diagrams, is an important step in enhancing the considerable modeling and reasoning capabilities offered by BNs.

In this paper, we develop a version of the *arc reversal* algorithm for networks with tree-structured CPTs. Arc reversal [16] is an important technique for manipulating BNs, and our approach demonstrates that structured CPTs can be exploited considerably. This allows smaller CPTs to be produced with less computational effort, and produces reversed networks that retain substantial structure in their CPTs; this structure can then be exploited in inference. In particular, the problems associated with increasing the number of parents a node has—a fact that makes reversal sometimes problematic—is mitigated by the use of structured CPTS.

We describe the relevance of our approach to stochastic simulation of *dynamic probabilistic networks* (DPNs) [5, 11, 10]. DPNs form an important class of BNs for modeling dynamical systems and sequential decision processes. Because of their size, exact methods are often rejected in favor of simulation techniques. In the case of DPNs, arc reversal or *evidence integration* [7] is extremely important; this case has been made forcefully [10]. However, even partial evidence integration can cause a large blowup in the size of CPTs; hence structured arc reversal can play an important role. We also show how the reversed DPNs can exploit the structured CPTs in simulation through the detection of irrelevance of variables *dynamically*. Specifically, we provide an algorithm that produces a *sampling schedule* for the variables within a "slice" of the DPN that ignores variables that can have no impact on the specific variables of interest. The process is dynamic in that the relevance of a certain variable to a query can depend on the context fixed by the earlier instantiation of other variables in a particular simulation trial; the CPTs allow one to identify the appropriate contexts. The net effect is that all variables throughout the DPN need not be sampled in every trial. The algorithm itself is time-independent, requiring processing of variables within a single time slice.

In Section 2, we describe CSI and the particular tree-structured CPTs we exploit in this paper. In Section 3, we develop the *tree-structured arc reversal* algorithm (TSAR). In Section 4, we describe the application of TSAR to the simulation of DPNs and elaborate on its benefits.

## 2 Context-Specific Independence

We assume a finite set $U = \{X_1, \ldots, X_n\}$ of discrete random variables where each variable $X_i$ may take on values from a finite domain $val(X)$. We use capital letters (e.g., $X, Y, Z$) for variable names and lowercase letters (e.g., $x$, $y$, $z$) to denote specific values taken by those variables. Sets of variables are denoted by boldface capital letters (e.g., $\mathbf{X}, \mathbf{Y}, \mathbf{Z}$), and assignments of values to the variables in these sets will be denoted by boldface lowercase letters (e.g., $\mathbf{x}, \mathbf{y}, \mathbf{z}$). We use $val(\mathbf{X})$ in the obvious way.

Concise representation of a joint distribution $P$ over this set of variables is one of the aims of Bayesian networks. A *Bayesian network* is a directed acyclic graph in which nodes correspond to these variables and arcs represent direct probabilistic dependence relations among these variables. Specifically, the structure of a BN encodes the set of independence assumptions corresponding to the assertion that each node $X_i$ is independent of its non-descendants in the graph given its parents $\Pi(X_i)$. These assertions are *local* in that they refer specifically to a node and its parents in the graph. Additional conditional independence relations of a more *global* nature can be determined efficiently using the graphical criterion of *d-separation* [13]. To represent the distribution $P$, we need only, in addition to the graph, specify for each variable $X_i$, a *conditional probability table (CPT)* encoding $P(\mathbf{x}_i \mid \Pi(X_i))$ for each possible value of the variables in $\{X_i, \Pi(X_i)\}$. (See [13] for details.)

Apart from the usual strong independence relations encoded in BNs, we are often interested in independence between variables that holds only in certain contexts. Let $\mathbf{X}, \mathbf{Y}, \mathbf{Z}, \mathbf{C}$ be pairwise disjoint sets of variables. We say $\mathbf{X}$ and $\mathbf{Y}$ are *contextually independent* [2] given $\mathbf{Z}$ and the *context* $\mathbf{c} \in val(\mathbf{C})$ if

$$P(\mathbf{X} \mid \mathbf{Z}, \mathbf{c}, \mathbf{Y}) = P(\mathbf{X} \mid \mathbf{Z}, \mathbf{c}) \text{ whenever } P(\mathbf{Y}, \mathbf{Z}, \mathbf{c}) > 0.$$

Thus, the independence relation between $\mathbf{X}$ and $\mathbf{Y}$ need *not* hold for all values $val(\mathbf{C})$.

Local statements of *context-specific independence (CSI)* can be detected in the CPTs for a node. For example, consider the CPT for variable $A$ illustrated in Figure 2. While $P(A)$ is directly dependent on variables $A', B', C', D'$, given the specific value $a'$ of $A'$, $A$ is not dependent on variables $B', C', D'$; that is, $P(A|a', B', C', D') = P(A|a')$. The structure inherent in the CPT is exploited in the decision-tree representation given in the figure (by convention, we take left arcs in trees to be labeled true, and right arcs false). In this example, the CPT for $A$ is encoded with 5 distinct entries rather than the 16 required by the usual tabular representation. We note that simple extensions of d-separation can be used to find global CSI relations [2].

It is suggested in [2] that CPTs can be encoded using appropriate compact function representations that make explicit such local CSI relations. We will focus on decision trees in this paper. We do not delve further into the details of CSI or the use of tree-structured CPTs in general (see [2] for further details). We do note that tree-structured CPTs and CSI have been exploited in decision-making [1], knowledge acquisition [9] and learning [6]. The integration of CSI with other well-known BN methods promises to make it even more pervasive. We now consider one such combination of tree-structured CPTs with a BN manipulation algorithm.

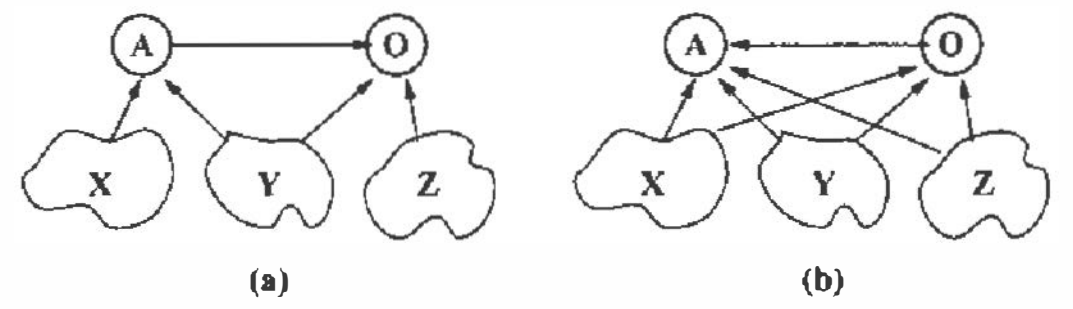

Figure 1: General Structure of Arc Reversal (Shachter)

## 3 Tree-Structured Arc Reversal

### 3.1 Arc Reversal with Unstructured CPTs

Arc reversal [16] is a technique for restructuring a BN so that the arc between two nodes has its directionality reversed, while still correctly representing the original distribution. Arc reversal is an important technique for BNs and influence diagrams, and plays a significant role in the evaluation of BNs through stochastic simulation [7, 10], as we describe in the next section.

The basic arc reversal operation is relatively straightforward. Consider a network where variable $A$ is a parent of $O$. The variables belonging to the set $\Pi(A) \cup \Pi(O)$ can be divided into three classes: $\mathbf{X} = \Pi(A) \setminus \Pi(O)$, $\mathbf{Y} = \Pi(A) \cap \Pi(O)$, and $\mathbf{Z} = \Pi(O) \setminus \Pi(A) \setminus \{A\}$ (see Figure 1(a)). Suppose we wish to reverse the arc between $A$ and $O$. To ensure that the resulting network makes correct independence assertions, we must add parents to both nodes $A$ and $O$: in particular, each element of $\mathbf{X}$ becomes a parent of $O$, and each element of $\mathbf{Z}$ becomes a parent of $A$. $A$ and $O$ retain their original parents as well (apart from the reversal of the arc between them). The structure of resulting network is illustrated in Figure 1(b). We use the notations $\Pi_{old}(A)$ and $\Pi_{new}(A)$ to refer to $A$'s parents before and after reversal, respectively (similarly for $O$).

The expressions for the new CPT entries are [16]:

$$P(O|\mathbf{x}, \mathbf{y}, \mathbf{z}) = \sum_{a \in val(A)} P(O|\mathbf{y}, \mathbf{z}, a) P(a|\mathbf{x}, \mathbf{y}) \quad (1)$$

$$P(A|\mathbf{x}, \mathbf{y}, \mathbf{z}, O) = \frac{P(O|A, \mathbf{y}, \mathbf{z}) P(A|\mathbf{x}, \mathbf{y})}{P(O|\mathbf{x}, \mathbf{y}, \mathbf{z})} \quad (2)$$

Note that each term in Equation 1 is in an original CPT, as are the terms in the numerator of Equation 2, while the denominator is simply an entry in the new CPT for $O$.

### 3.2 Arc Reversal with Structured CPTs

Arc reversal often significantly increases the number of parents of the nodes $A$ and $O$ involved. Since CPT size increases exponentially with the number of parents, the resulting CPTs can become very large and require a prohibitive amount of computation to construct. However, should the CPTs in the original network exhibit structural regularities, one would expect the new CPTs to do the same. The key to preserving structure during arc reversal is to be able to identify, using only the structure of the original CPTs, the regularities in the new CPT. The net result is a smaller structured CPT for the nodes involved in the reversal, as well as the benefit of the computational savings associated with computing fewer (distinct) CPT entries. In addition, inference algorithms that exploit CSI can be used after reversal if we are able to retain this structure.[1]

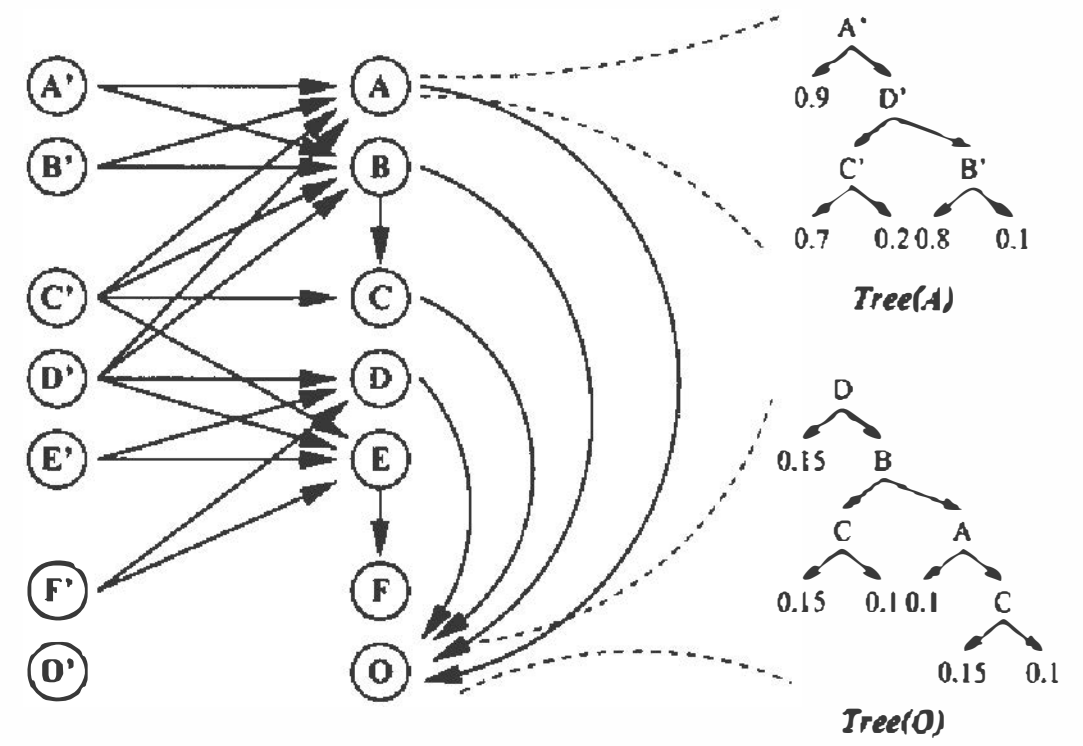

Figure 2: A Slice of a DPN

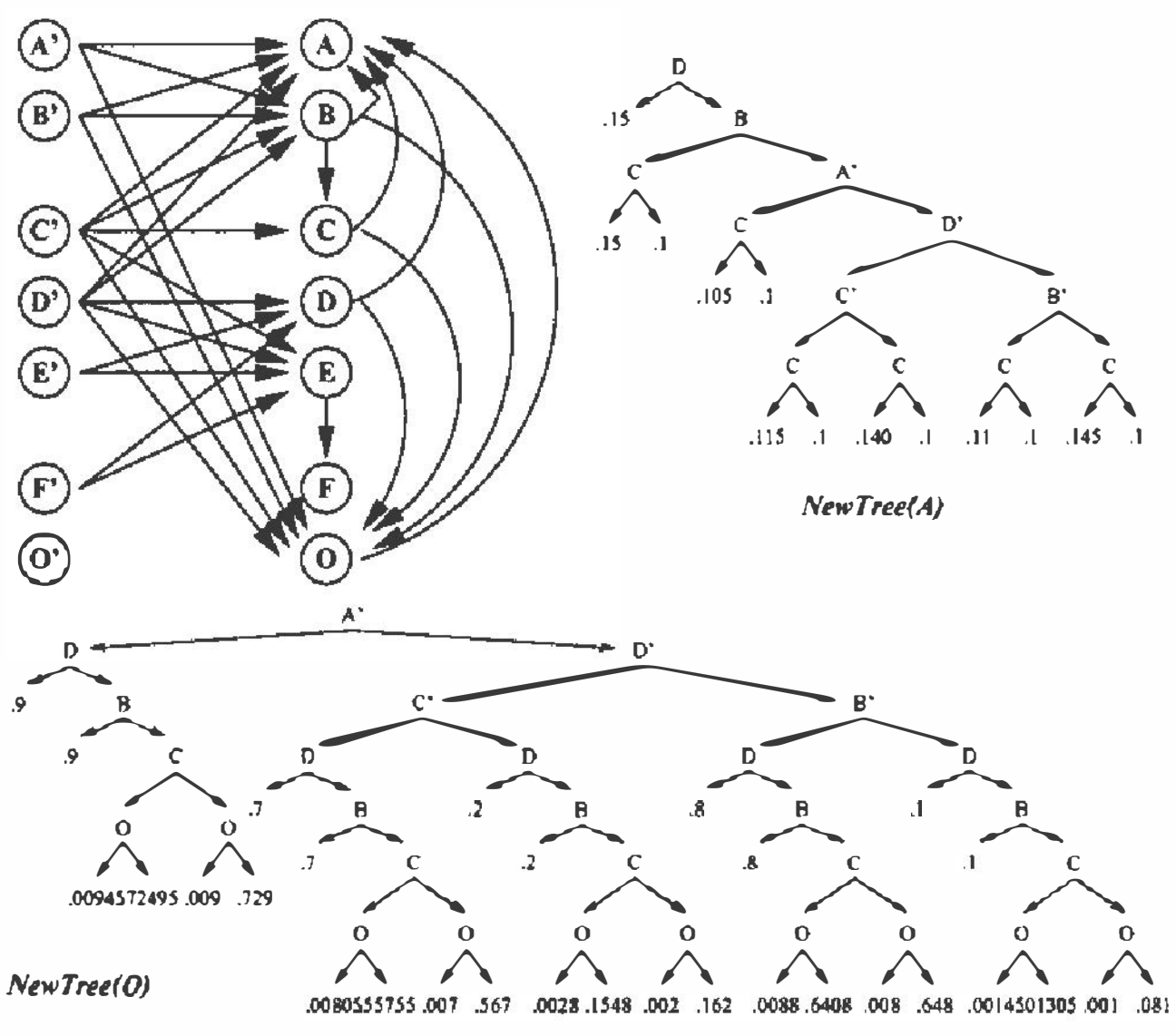

Figure 3: DPN after Reversal between $A$ and $O$

We now describe a simple *tree-structured arc reversal* (TSAR) algorithm for constructing tree-structured CPTs for nodes involved in arc reversal assuming tree-structured CPTs for the original network.

We use the network shown in Figure 2 to illustrate the basic intuitions underlying TSAR.[2] We consider the reversal of the arc between $A$ and $O$, where the tree-structured CPTs for both variables are shown in the figure. For ease of exposition, all variables in the example are boolean.

When reversing the arc from $A$ to $O$, we have $\Pi_{new}(A) = \{A', B', C', D', B, C, D, O\}$ and $\Pi_{new}(O) = \{B, C, D, A', B', C', D'\}$, as indicated in Figure 3. Had the CPTs for this network been represented in an unstructured form, standard arc reversal would produce a new CPT for $A$ with $2^8 = 256$ entries and a new CPT for $O$ with $2^7 = 128$ entries, and would have required a proportional amount of computation. However, it is clear that since many of the original CPT entries for the nodes $A$ and $O$ are identical for different assignments to their parents, the new CPTs must also have many redundancies.

Consider first the new CPT for $O$. The following local CSI relations hold between $O$ and its new parents:[3]

- Let $\boldsymbol{y}', \boldsymbol{z}'$ be some instantiation of $O$'s original parents (i.e., $\boldsymbol{Y}' \subseteq \boldsymbol{Y}$, $\boldsymbol{Z}' \subseteq \boldsymbol{Z}$), such that $O$ is independent of $A$ given $\boldsymbol{y}', \boldsymbol{z}'$. Then for any instantiation $\boldsymbol{x}, \boldsymbol{y}, \boldsymbol{z}$ of $O$'s new parents consistent with $\boldsymbol{y}', \boldsymbol{z}'$, we have (by Equation 1) that $P(O|\boldsymbol{x}, \boldsymbol{y}, \boldsymbol{z}) = P(O|\boldsymbol{y}', \boldsymbol{z}')$. For example, $O$ is independent of $A$ given $d$ (see Figure 2), so $P(O|d)$ remains unchanged for any assignment to $\Pi_{new}(O)$.
- Let $V$ be some variable in $\boldsymbol{X} = \Pi(A) \setminus \Pi(O)$. If, for some instantiation $\boldsymbol{x}', \boldsymbol{y}'$ of a subset of $A$'s original parents (i.e., $\boldsymbol{X}' \subseteq \boldsymbol{X}, \boldsymbol{Y}' \subseteq \boldsymbol{Y}$), $A$ is independent of $V$, then for any instantiation of a similar subset of $O$'s new parents of the form $\boldsymbol{x}', \boldsymbol{y}', \boldsymbol{z}$, $O$ is independent of $V$. For example, the original tree for $A$ indicates that $A$ and $D'$ are independent given $a'$. Since $D'$ is not an original parent of $O$, the probability of $O$ given its new parents cannot vary with $D'$ given $a'$.
- Let $V$ be some variable in $\boldsymbol{Z} = \Pi(O) \setminus \Pi(A) \setminus \{A\}$ such that, for some instantiation $\boldsymbol{y}', \boldsymbol{z}'$ of a subset of $O$'s original parents (i.e., $\boldsymbol{Y}' \subseteq \boldsymbol{Y}$, $\boldsymbol{Z}' \subseteq \boldsymbol{Z}$), $O$ is independent of $V$ given $\boldsymbol{y}', \boldsymbol{z}', a_i$ for all values $a_i$ of $A$. Then for any instantiation of a similar subset of $O$'s new parents of the form $\boldsymbol{x}, \boldsymbol{y}', \boldsymbol{z}'$, $O$ remains independent of $V$.

These three observations give rise to a simple algorithm for construction a CPT-tree for $O$ given its new parents, where an arc from $A$ to $O$ is being reversed. The algorithm proceeds as follows:

1. Create a copy of $Tree_{old}(O)$, removing any subtrees that lie below a node labeled $A$, resulting in (the initial component of) $Tree_{new}(O)$. For each $A$-node (say $A_j$) in the tree, record the subtrees associated with each value $a \in val(A)$; we denote these trees by $Tree_j(a)$.

[1] We expect these ideas to be applicable to compact CPT representations apart from trees.

[2] This network represents a "slice" of a DPN (see next section).

[3] We use sets $\boldsymbol{X}, \boldsymbol{Y}, \boldsymbol{Z}$ as in Figure 1; in this example, $\boldsymbol{X} = \{A', B', C', D'\}$, $\boldsymbol{Y} = \emptyset$, and $\boldsymbol{Z} = \{B, C, D\}$.

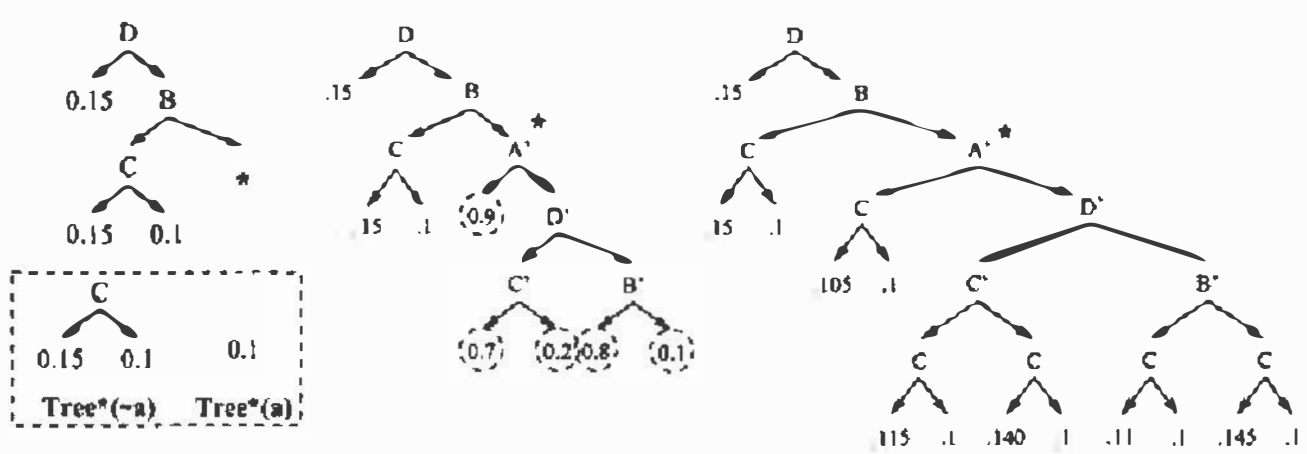

Figure 4: Construction of CPT Tree for $O$

2. For each $A$-node in $Tree_{new}(O)$ (which must be a leaf), *Graft* a copy of $Tree_{old}(A)$ onto this location. *Reduce* the copy of $Tree_{old}(A)$ by deleting any redundant nodes. We denote by $Tree_j(A)$ the copy of $Tree_{old}(A)$ added to location $A_j$. We also *mark* the root of $Tree_j(A)$.
3. For each $A_j$:
   (a) *Merge* the trees $\{Tree_j(a) : a \in val(A)\}$, recording the values $P(O|a)$ for each $a \in val(A)$ at the leaves; denote the result by $Tree_j(O|A)$.
   (b) Graft a copy of $Tree_j(O|A)$ to each leaf of $Tree_j(A)$.
   (c) For each copy so added, compute the value of Equation 1 using the terms $P(O|a)$ recorded at the leaves of $Tree_j(O|A)$ and the $P(a)$ terms recorded at the leaf of $Tree_j(A)$ to which the copy was grafted.

We elaborate on the details by referring to the running example. Step 1 requires that we duplicate all of $Tree_{old}(O)$ except for subtrees that lie under any node labeled with variable $A$. This is shown in Figure 4(a), where the asterisk denotes the location where the removal of the $A$ subtrees occurred (these are recorded below). Any complete branch of that remains denotes a context in which $O$ is independent of $A$, and thus independent of *all other new parents*: the probability at the leaf node is unchanged. In Figure 4(a), we see that no computation is needed to determine $P(O|d)$, $P(O|\overline{d}bc)$ or $P(O|\overline{d}b\overline{c})$ in the new CPT.

Step 2 involves the replacement of any subtree whose root is labeled with variable $A$ by $Tree_{old}(A)$. This is necessary because $P_{new}(O)$ depends on the probability of $A$ given its old parents. This is illustrated in Figure 4(b) where $Tree_{old}(A)$ is "grafted" to $Tree_{old}(O)$ where node $A$ was located.[4] At each of the leaf nodes (circled in the example), we now have $P(A)$ given its old parents recorded. While not applicable in our example, Step 2 performs *tree reduction* as well, removing any redundant nodes that may have been added. If, for example, $D$ were a parent of $A$, all occurrences of $D$ would be removed from $Tree_{old}(A)$ before grafting, since the value of $D$ is fixed to $\overline{d}$ earlier in the tree (from $Tree_{old}(O)$). Any node labeled $D$ would be replaced by the appropriate $\overline{d}$ subtree. This can play a substantial role if $A$ and $O$ share parents (they share none in our example). Finally, the diagram shows that the root of this grafted tree is marked with an asterisk. This notes the fact that $P(O)$ at the leaves of this subtree may in fact be different from their values in $Tree_{old}(O)$, while any leaf that does not lie below a marked node is such that $P(O)$ is identical to its value in $Tree_{old}(O)$. Such marks are used below in the construction of $Tree_{new}(A)$, as described later.

[4] In general, this takes place at every occurrence of node $A$.

Step 3 of the algorithm requires that the subtrees corresponding to *each* value of $A$ that were removed from $Tree_{old}(O)$ at that point now be grafted onto each leaf of $Tree_{old}(A)$ that was just added. More precisely, for each such node $A$ that was replaced in $Tree_{old}(O)$, we *merge* each of its subtrees.[5] The leaves of this merged tree dictate $P(O|a)$ for each $a \in val(A)$ (given the other relevant parents). The merged tree is then grafted onto the leaves of the relevant copy of $Tree_{old}(A)$. At the leaf of any "copy" of the merged tree, we compute the value of Equation 1, with the required terms readily available. Figure 4(c) illustrates this process. For the (single) copy of $Tree_{old}(A)$ that has been added, we merge the subtrees $a$ and $\overline{a}$ subtrees that were removed from $Tree_{old}(O)$ at that point: since the $a$ subtree is empty, the merged tree is simply the $\overline{a}$ subtree. We record both $P(O|a)$ and $P(O|\overline{a})$ at each leaf node of the merged tree. This tree is then copied to each of the five leaf nodes where $P(a)$ is recorded; and the required conditioning computation takes place for each resulting leaf node.

The CSI reflected in the resulting $Tree_{new}(O)$ is sound:

**Theorem 1** *Let c be some context determined by a branch of $Tree_{new}(O)$ instantiating variables $\boldsymbol{C} \subseteq \Pi_{new}(O)$. Then $O$ is contextually independent of $\Pi_{new}(O) \setminus \boldsymbol{C}$ given $c$.*

The computational and space savings can be considerable when constructing the new CPT for $O$ in tree form. As mentioned, the use of tabular CPTs would produce a new CPT for $O$ with 128 entries and require 128 calculations of Equation 1. In this example, much of the tree structure of the original CPTs is preserved in $Tree_{new}(O)$: it requires only 13 distinct CPT entries and only 10 calculations of Equation 1 (since 3 of the entries are retained from $Tree_{old}(O)$).

We now turn our attention to the construction of $Tree_{new}(A)$ via Equation 2. Again, certain CSI relations hold:

- Let $\boldsymbol{x}', \boldsymbol{y}', \boldsymbol{z}'$ be some partial instantiation of $O$'s new parents (i.e., $\boldsymbol{X}' \subseteq \boldsymbol{X}, \boldsymbol{Y}' \subseteq \boldsymbol{Y}, \boldsymbol{Z}' \subseteq \boldsymbol{Z}$), such that $O$ is independent of its remaining new parents given $\boldsymbol{x}', \boldsymbol{y}', \boldsymbol{z}'$, and $A$ is independent of its remaining original parents (i.e., those in $\boldsymbol{X} \cup \boldsymbol{Y}$) given $\boldsymbol{x}', \boldsymbol{y}'$. Then $A$ is independent of its remaining *new* parents given $\boldsymbol{x}', \boldsymbol{y}', \boldsymbol{z}'$ and $O$ (by Equation 2). In our example, $A$ is independent of its other parents given $a'$ while $O$ is independent of (other) new parents given $\overline{d}ba'c$. Thus, $A$ is independent of $B'$, $C'$ and $D'$ given $\overline{d}ba'c$ and $O$.
- Let $\boldsymbol{x}', \boldsymbol{y}', \boldsymbol{z}'$ be some partial instantiation of $O$'s new parents such that $P(O|\boldsymbol{x}', \boldsymbol{y}', \boldsymbol{z}') = P(O|\boldsymbol{y}', \boldsymbol{z}')$; that is, $O$ is independent of $A$ given $\boldsymbol{y}', \boldsymbol{z}'$. Then, by Equation 2, $P(A|\boldsymbol{x}', \boldsymbol{y}', \boldsymbol{z}', O) = P(A|\boldsymbol{x}', \boldsymbol{y}')$. For example, $d$ renders $O$ independent of its parents in both $Tree_{old}(O)$ and $Tree_{new}(O)$; in particular, $O$ is independent of $A$. Thus, any instantiation of $A$'s old parents that fixes $P(A)$ (e.g., $a'$) determines $P(A)$ given its new parents. In our example $P(A|a'd) = P(A|a')$ and $A$ is independent of other new parents given $a'd$.

[5] Merging simply requires creating a tree whose branches make the distinction contained in each subtree. We do this by ordering the trees, and grafting each tree in order onto the leaves of the tree resulting from merging its predecessor, removing redundant nodes as appropriate.

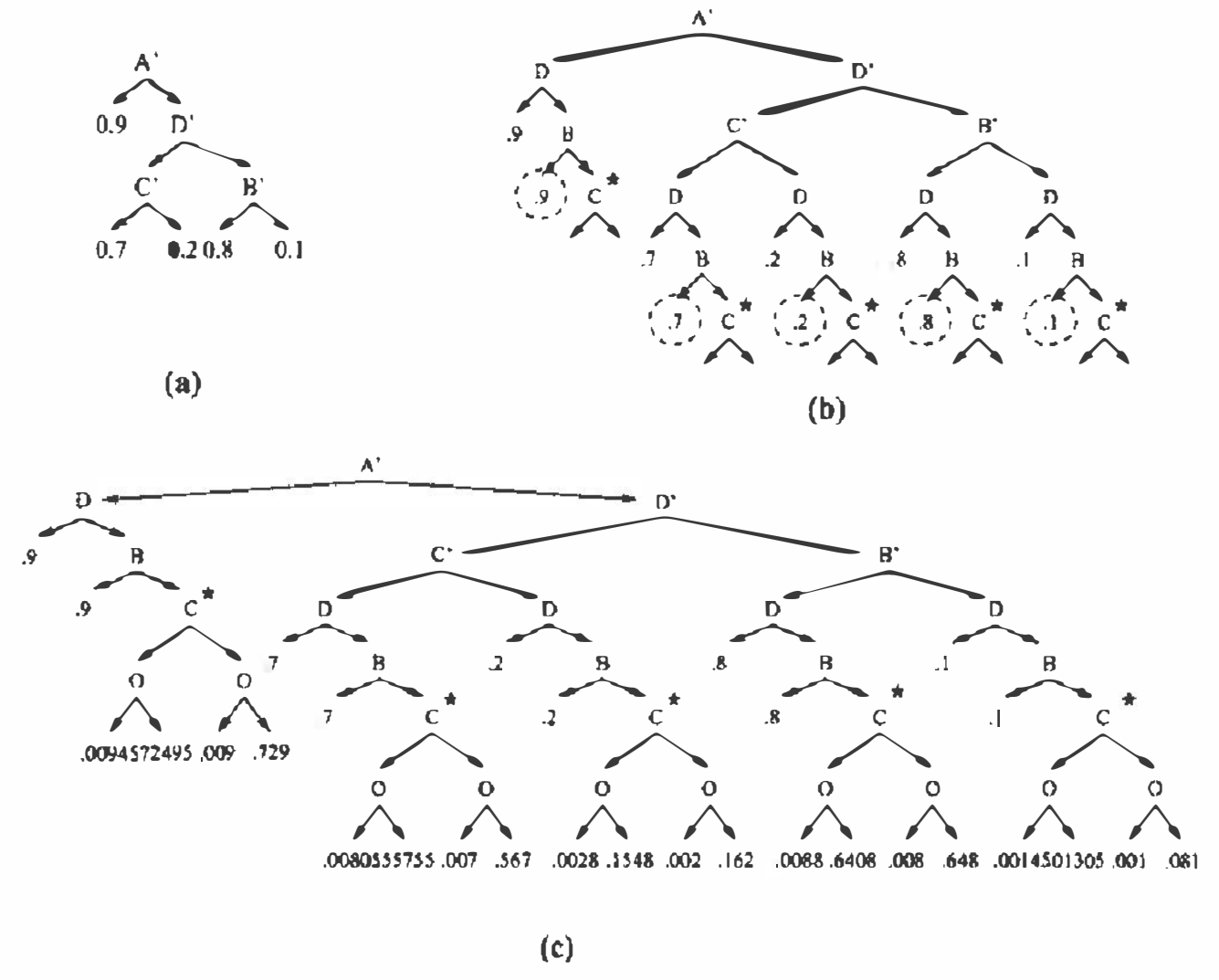

Figure 5: Construction of CPT Tree for $A$

These two observations give rise to a simple algorithm for constructing a CPT-tree for $A$ given its new parents, where an arc from $A$ to $O$ is being reversed.

1. Create a copy of $Tree_{old}(A)$.
2. For each leaf $l$ of $Tree_{old}(A)$:
   (a) Graft a copy of $Tree_{new}(O)$ to the leaf, and reduce the tree by removing redundant nodes (record the distribution $P_l(A)$ labeling the leaf $l$).
   (b) *Collapse* any subtree of the reduced $Tree_{new}(O)$ which is not marked as altered into a single leaf node; denote this reduced, collapsed tree $Tree_l(O)$.
   (c) Label each unmarked leaf of $Tree_l(O)$ with $P_l(A)$.
   (d) Add a node $O$ to each leaf $lo$ in the marked subtrees of $Tree_l(O)$ (and record the distribution $P_{lo}(O)$ labeling leaf $lo$).
   (e) For each new leaf under the $O$-nodes, compute the value of Equation 2 using the terms $P_l(A)$, $P_{lo}(O)$, and the values $P(O|\Pi_{old}(O))$ determined from $Tree_{old}(O)$.

We elaborate on the details by referring to the running example. Step 1 requires that we duplicate all of $Tree_{old}(A)$ and keep track of the distribution labeling each leaf. This is shown in Figure 5(a). Step 2 involves a number of substeps. First, a copy of $Tree_{new}(O)$ is grafted to each leaf of $Tree_{old}(A)$ and reduced as shown in Figure 5(b). It also shows how the left subtree under each $B$ node is collapsed by removal of the $C$ node (the circled leaves), and how each "unmarked" leaf inherits $P(A)$ from $Tree_{old}(A)$: since $P(O)$ is identical at unmarked leaves given $\Pi_{new}(O)$ or $\Pi_{old}(O)$, these terms in Equation 2 cancel (thus $P(A)$ need not be computed). Finally, Figure 5(c) shows the addition of the node for new parent $O$ at marked leaf, and the values of $P(A)$ computed according to Equation 2.

Again, the resulting $Tree_{new}(A)$ is sound:

**Theorem 2** *Let* $c$ *be a context determined by a branch of* $Tree_{new}(A)$ *instantiating variables* $\boldsymbol{C} \subseteq \Pi_{new}(A)$. *Then* $A$ *is contextually independent of* $\Pi_{new}(A) \setminus \boldsymbol{C}$ *given* $c$.

Once again we note that this algorithm preserves a considerable amount of structure in this example. Using unstructured CPTs, the new CPT for $A$ would require $2^8 = 256$ entries and the same number of evaluations of Equation 2. Exploitation of tree-structured CPTs allows the new CPT for $A$ to be expressed with only 30 distinct entries, and requires that Equation 2 be evaluated only 20 times (i.e., at the leaves of marked subtrees).[6]

## 4 TSAR in the Simulation of DPNs

Dynamic probabilistic networks (DPNs) are a particular form of BN used to model temporally-extended systems [5, 11, 10]. Intuitively, we imagine a number of state variables whose values vary over time, allowing the network to be organized in "slices" consisting of a set of variables at a particular point in time. The system dynamics are often taken to be Markovian and stationary: the causal influences for any variable at time $t$ must be drawn from the set of variables at time $t-1$ or time $t$, and this relation holds for all times $t$. Thus the DPN can be represented schematically in a very compact fashion by simply representing the relation between two consecutive generic slices at time $t$ and $t+1$ (together with priors for root nodes at time 1).

DPNs can be used to model dynamical systems generally, and specifically can be applied to time series models [11], control problems such as robot or vehicle monitoring and control [12, 10], planning [5] and sequential decision problems [18, 1]. We often distinguish certain variables within a particular slice as *state variables*, and others as *sensor variables*. It is generally only sensor variables that are observable and provide evidence of the system's trajectory. This is illustrated schematically in Figure 6 (following [10]). We note that the set of state and sensor variables need not be disjoint, and that state variables could include *decision variables* whose values are set by the controller (possibly depending on the values of previous state variables). Figure 2 (used earlier) illustrates a DPN with a number of state variables ($A$ through $F$) and a single sensor variable ($O$). Our convention is that node $A'$ denotes variable $A_t$ ($A$ at time $t$) and node $A$ denotes $A_{t+1}$.

[6] We note that these 30 entries are, in fact, not all distinct. But the tree-representation imposes certain redundancies that could be overcome using other function representations.

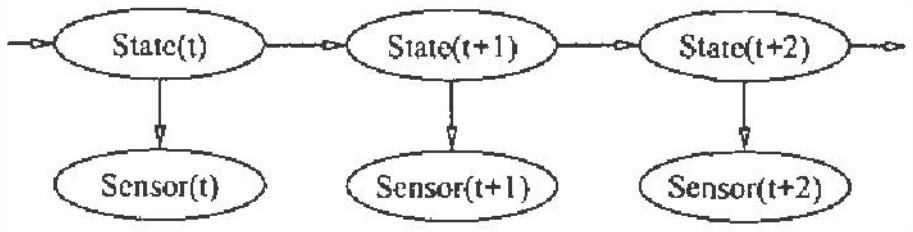

Figure 6: Schematic Representation of a DPN

A common task in DPNs is *projection* or *forecasting*, that is, determining, at time $t$, the distribution over some subset of future variables (i.e., state variables at times later than $t$) given a set of observations at some points in the past (i.e., evidence at sensor variables at certain points prior to time $t$). For example, one might want to compute the expected value of a particular policy for some $k$ steps into the future given observations of past behavior of the system. Because of the size of DPNs, exact solution of a DPN is impractical in most settings. Therefore simulation models are often preferred. However, traditional methods such as *likelihood weighting* [17, 7] will be extremely unsuitable in DPNs exhibiting the schematic structure of Figure 6. Because they are sinks in the network, the sensor variables (which provide the only evidence) are unable to influence the course of the simulation. As demonstrated convincingly by Kanazawa, Koller and Russell [10], straightforward simulation will often "get off track" very quickly, leading to trials with negligible (or zero) weight. They suggest the use of (partial) evidence integration in order to keep the simulation close to reality. Intuitively, arcs from state variables to sensors within time $t$ are reversed so that observed evidence will strongly influence the sampled state at time $t$. The reversal is only partial, however, since sensor variables in the reversed network will generally have as parents state variables from slice $t-1$.[7]

Unfortunately, evidence integration can be expensive. Indeed, Fung and Chang [7] suggest that, while evidence integration can help convergence of simulation methods tremendously, the computational cost of arc reversal may prove to be a practical obstacle to its applicability. Fortunately, in DPNs, evidence integration benefits from the uniform nature of the network: the reversal needs to be computed at one slice only, and can then be applied across all slices. Thus, one substantial burden is overcome. Yet complete "within slice" integration of sensor variables can still be rather costly. Consider the network in Figure 2. Complete reversal of arcs into sensor variable $O$ (within a single slice only) results in the extremely connected network illustrated in Figure 7. Variables $A$, $B$, $C$, $D$ and $O$ have CPTs of sizes 256, 128, 64, 128 and 64, respectively. Furthermore, $O$ (which is involved in four reversals) has intermediate CPTs of sizes 128, 64 and 32. Arc reversal requires explicit computation of each of these 864 entries. In larger networks, with tens or hundreds of state variables, this can prove a major impediment to evidence integration.

[7] We take as accepted the crucial role of evidence integration in the convergence of simulation involving DPNs. Our experiments with networks of this type (both with tree-structured CPTs and without) confirm this impression.

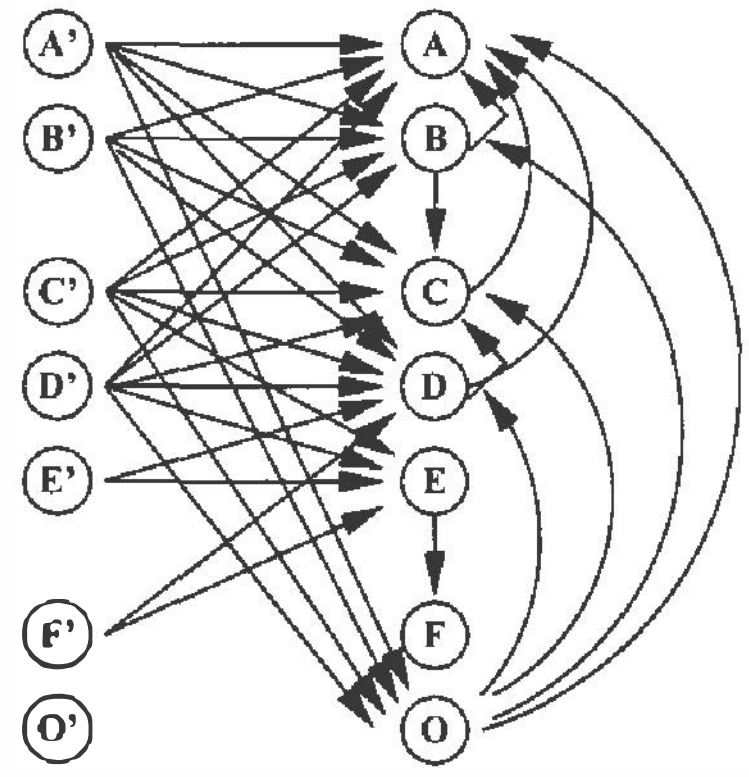

Figure 7: Reversal of $O$ and all "In-Slice" Parents

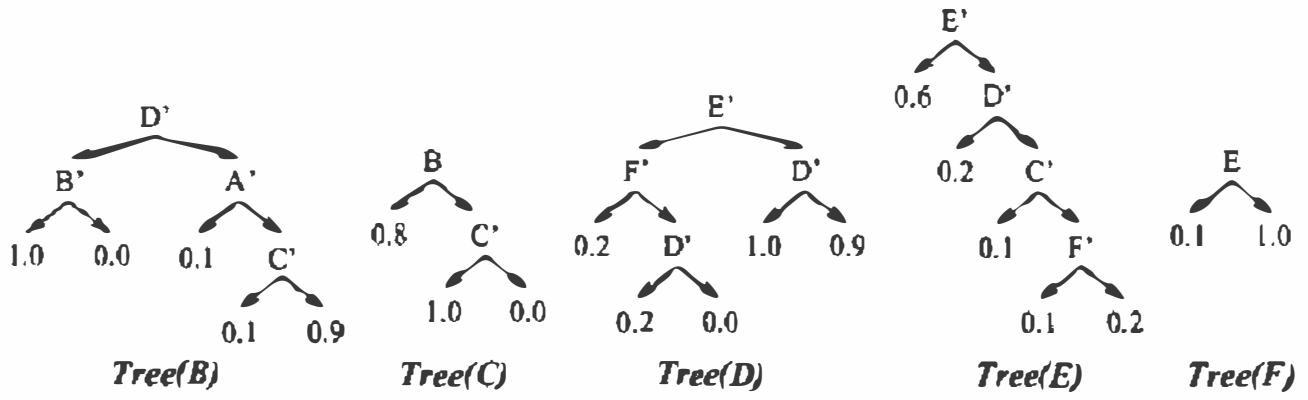

Figure 8: CPTs for Remaining Variables

This suggests that tree-structured arc reversal can play an important role in the simulation of DPNs. The computational burden of evidence integration as well as the size of the resulting CPTs can be considerably lessened by the use of TSAR. For instance, if the CPTs of the remaining nodes in our example DPN are as shown in Figure 8, the sizes of the trees in the reversed network for $A$, $B$, $C$, $D$ and $O$ are 30, 33, 23, 72 and 36, respectively, while $O$'s intermediate trees have sizes 13, 11 and 15. Of the total 233 CPT entries represented, only 210 required explicit computation. Our experiments with other similar DPNs suggest that this savings is commonplace. This seems especially true in the evaluation of policies, where *actions* or decisions play a predominant role. As argued in [3], the representation of action effects often admits a considerable amount of CSI.

Apart from the potential savings it provides during network restructuring, another advantage offered by TSAR is the ability to determine *irrelevant* variables *dynamically*. Irrelevance can be viewed at the network level. For instance, in the DPN above, we may be interested in the distribution of variable $A_t$. Given such a specific query, a simulation trial need not sample $F_{t-1}$ since this cannot impact $A_t$. The *irrelevance* of $F_{t-1}$ to $A_t$ is dictated by the structure of the DPN. Indeed, Fung and Chang [7] propose irrelevance of this type as a means of speeding simulation, though they caution that the overhead involved may offset any savings.[8]

[8] See also [14] for a discussion of this type of relevance.

We focus on a specific problem: assume a DPN has been given and that a certain subset of state variables has been designated as *immediately relevant*. The simulation is designed to sample these variables over time; the fact that other variables are being sampled is subsidiary to this aim.[9] In our example, imagine that $A$ is the variable of interest. What we wish to determine is the set of variables that must be sampled in each slice to ensure we can accurately determine the conditional probability of $A$ at any future time.

We would hope that "schematic" detection of irrelevance could alleviate overhead difficulties (i.e., the detection of irrelevance by processing a single slice in a manner that applies across all time points). But, clearly, if we need to sample $A$ at each slice, we cannot ignore $F$: while $F_{t-1}$ doesn't impact $A_t$, it does influence $A_{t+1}$ through its impact on $D_t$. The influence of certain variables often "bleeds through" to many or all other variables over time, making network-level irrelevance unhelpful. Fortunately, the tree-structured CPTs suggest that some variables may be irrelevant to others *under certain conditions*, even if they are not irrelevant at all times. So, while $F_{t-1}$ might be required in order to sample $D_t$ (which itself impacts $A_{t+1}$, the variable of interest), the tree for $D$ shows that $F_{t-1}$ has no impact if $E_{t-1}$ is true. This suggests that one should sample $E$ at a particular time slice first; and if it $E$ turns out to be true, one should not sample $F$ at that slice. Some care is required of course, since $F$ may influence other variables of relevance through a different causal chain.

Our goal is a simple algorithm for constructing a *conditional sample schedule* for the variables within a time slice in which a variable is not sampled if it provably has no influence on *any* variable of interest at *any* future point in time. An example of constraints on such sample schedule might be: "Generate values for $E, D, C$ before $F$. If $\overline{e}d$ or $\overline{edc}$, do not sample $F$." Our algorithm proceeds in four phases.

We first identify *(unconditionally) relevant variables*, those variables that can influence the future value of some variable of interest; in our example, all variables are relevant since all influence $A$ (the variable of interest) over time. This can easily be detected using the topological structure of the network. For our purposes, we now treat the set of unconditionally relevant variables as potentially relevant to the future values of immediately relevant variables.

Second, we construct a *sample graph* for each relevant variable. This structure describes the dependence of each variable on other variables within the same or previous slice. Essentially, these are directed acyclic graphs generated from the CPTs for the variables in question, and are similar to binary decision diagrams (BDDs) [4]. The sample graphs for the seven variables are shown in Figure 9. For example, the graph for variable $O$ dictates that, in order to sample it: we need the value of $E'$; if $\overline{e}'$, we need $F$, otherwise we proceed to $D$; once we sample $F$ (if necessary) we proceed to $D$ and so on.[10]

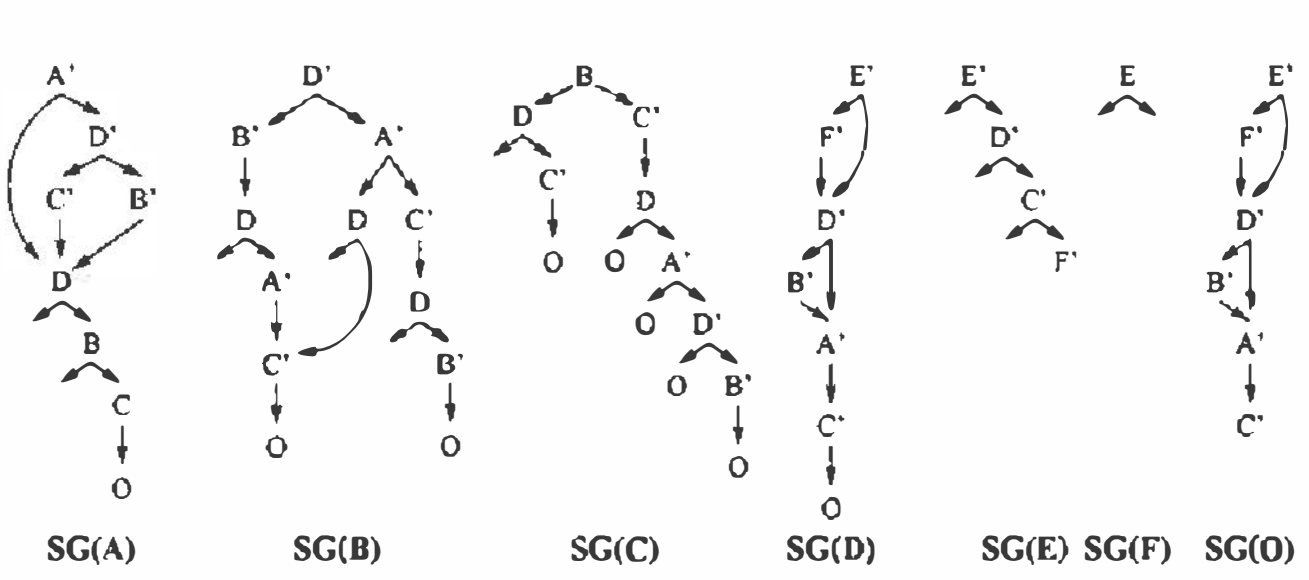

Figure 9: Sample Graphs for all Variables

The key third phase of the algorithm requires that we determine the conditions under which a (unconditionally relevant) variable is *conditionally irrelevant*. To begin, we order variables so that the variables first in the ordering have sample graphs (or CPTs) that depend only on previous variables, and that variables later in the ordering depend only on variables in the same time slice that lie earlier in this ordering. (A suitable ordering for this example is $O, D, B, E, F, C, A$.) Then for each variable, we determine the conditions under which it is *required* by unconditionally relevant variables in the next slice (i.e., when must it be known in order to determine the distribution for that variable). Using the ordering of variables suggested, we apply the variable in question to each sample graph, determining the set of minimal initial segments of paths (or contexts) in the graph that have no completion leading to the variable in question.

To illustrate, we consider the conditions under which the value of $F$ at one slice is *not required* to accurately sample other variables at the next (or any future) time slice. We *apply* $F$ to each sample graph, in turn, using our variable ordering. applying $F$ to the graph for $O$, we see that condition $\overline{e}$ is the only one that guarantees $F$'s value is not required to sample $O$. If we apply $F$ to the graph for $D$, we see again that $\overline{e}$ is the only such condition. If we then process the graph for $B$, we see that $F$ does not occur, but $B$ depends on the current value of $D$; since we have already processed $D$'s graph, at that point we can insert the discovered condition for $D$ (i.e., $\overline{e}$) at that point in our search through $B$'s graph (thus, the ordering of variables plays an important role). Note that if several distinct paths bypass $F$, the condition generated is disjunctive: in applying $F$ to $E$' sample graph, we see that $F$ is irrelevant to $E$ if $e \vee \overline{e}d \vee \overline{edc}$. We note that $F$ is irrelevant when $\overline{e}$ for all other variables.[11]

[9] For instance, in policy evaluation, one may wish simply to sample and sum value nodes at each slice, with no "direct" interest in other state variables being expressed.

[10] Intuitively, such a graph can be constructed by joining common subtrees (ignoring leaf values) in the CPT, and collapsing true/false branches from a variables that lead to similar subtrees. This graph can easily be built while the tree is being constructed during arc reversal (if the node is part of a reversed arc). The complexity of identifying common subtrees should not be a limiting factor in this setting. But we should point out that the ideas below can be applied using the trees themselves. Collapsing trees into graphs simply aids the process somewhat.

[11] We note that finding all paths and other operations on sam-

Once we have determined the conditions under which $F$ is irrelevant for all variables, their conjunction fixes the conditions under which $F$ is not needed to determine *any value at the next slice*: in this case, we obtain $\bar{e}d \vee \overline{edc}$. We note that variables other than $F$ in our example can be ignored under any conditions.

The conditions so obtained for $F$ suggest that one should sample variables within any given slice such that $F$ is sampled after $E$, $D$ and $C$, and only when $e$ or $\overline{edc}$ obtains. Of course, the conditions obtained for other variables might impose other contraints on the ordering. Phase 4 of the algorithm involves construction of a *sample schedule* that satisfies as many constraints as possible. In this example, since no other variables are irrelevant under any conditions, we simply use this schedule. One could imagine, however, that one might want to sample $F$ before $E$ because of their impact on a third variable. In such a case one could not satisfy both the requirement to sample $F$ before $E$ and the requirement to sample $E$ before $F$. In this case, an arbitrary choice could be made, or some heuristic could be used (e.g., if we had some estimate of the steady state probabilities that suggested $\bar{e}$ was unlikely, we would know that skipping $F$ was also unlikely, in which case, we might decide to opt for conditionally sampling $E$ based on $F$ rather than vice versa).

Our running example was not designed to ensure a lot of conditional irrelevance, but it does offer the ability to not sample one variable (in each slice) under some conditions. For instance, a simple experiment using $A$ as the immediately relevant variable shows that $F$ needs to be sampled in only about a third of the slices: using 20 randomly generated observation sets (of 20 observations each) for our network stretched over 100 time slices, we saw that $F$ was sampled an average of 35 times out of the possible 100 times per run (the results were averaged over 1000 runs per evidence set). While not a large savings in this case (it is only one variable), the savings is proportional to the horizon of interest. For larger networks with substantial horizons, one might generally expect considerable savings from irrelevance processing. The longer the horizon, the less significant is the overhead involved in the (single slice) processing required. Furthermore, we expect that in large networks, a few key *contexts* (rather than variables) may shield variables of interest from large parts of the network.

## 5 Concluding Remarks

We have described an algorithm for tree-structured arc reversal and demonstrated its potential significance for the simulation of DPNs. Advantages include the reduction in (space and computational) overhead for reversal, and the ability to exploit the structured nature of the resulting reversed DPNs, especially in dynamic irrelevance detection. There are a number of questions that remain to be addressed. These include validation of the potential gains offered by TSAR and its use in simulation in realistic networks, and the application of these ideas to other forms of structured BNs. The existence of benchmark DPNs would aid this study. We are also investigating the potential of these ideas in the evaluation of policies for Markov decision processes.

ple graphs can use some of the efficient procedures designed for BDD manipulation [4]. Furthermore, this process can be terminated early if we ever find that the irrelevant condition for a variable with respect to any graph is false, or if the conjunction of conditions for any (incremental) subset of the graphs is inconsistent: the variable must then be sampled no matter what. For instance, when processing variable $E$ (or $D, A, C$) on the graph for $O$, we see that it must always be sampled, in which case application to other graphs is pointless.

**Acknowledgements:** This research was supported by NSERC Research Grant OGP0121843.